\documentclass{article}
\usepackage[utf8]{inputenc}
\usepackage{geometry}\usepackage[justification=centering]{caption}
\usepackage{amsmath}
\usepackage[table, dvipsnames]{xcolor}
\usepackage{amssymb}

\usepackage{mathptmx}       
\usepackage{helvet}         
\usepackage{courier}        
\usepackage{type1cm}        
\usepackage{makeidx}         
\usepackage{graphicx}        
\usepackage{tabularx}
\usepackage{float}
\usepackage{subcaption}
\usepackage{multirow}
\usepackage{textcomp}
\usepackage[linesnumbered,ruled]{algorithm2e}
\usepackage{multicol}        
\usepackage[bottom]{footmisc}
\title{A Cyber Threat Intelligence Sharing Scheme based on Federated Learning for Network Intrusion Detection}
\author{Mohanad Sarhan, Siamak Layeghy, Nour Moustafa, Marius Portmann}

\begin{document}

\maketitle

\begin{abstract}
    The uses of Machine Learning (ML) technologies in the detection of network attacks have been proven to be effective when designed and evaluated in a single organisation. However, it has been very challenging to design an ML-based detection system by utilising heterogeneous network data samples originating from several sources. This is mainly due to privacy concerns and the lack of a universal format of datasets. In this paper, we propose a collaborative federated learning scheme to address these issues. The proposed framework allows multiple organisations to join forces in the design, training, and evaluation of a robust ML-based network intrusion detection system. The threat intelligence scheme utilises two critical aspects for its application; the availability of network data traffic in a common format to allow for the extraction of meaningful patterns across data sources. Secondly, the adoption of a federated learning mechanism to avoid the necessity of sharing sensitive users' information between organisations. As a result, each organisation benefits from other organisations cyber threat intelligence while maintaining the privacy of its data internally/ The model is trained locally and only the updated weights are shared with the remaining participants in the federated averaging process. The framework has been designed and evaluated in this paper by using two key datasets in a NetFlow format known as NF-UNSW-NB15-v2 and NF-BoT-IoT-v2. Two other common scenarios are considered in the evaluation process; a centralised training method where the local data samples are shared with other organisations and a localised training method where no threat intelligence is shared. The results demonstrate the efficiency and effectiveness of the proposed framework by designing a universal ML model effectively classifying benign and intrusive traffic originating from multiple organisations without the need for local data exchange.
\end{abstract}

\section{Introduction}

Network Intrusion Detection Systems (NIDS) are tools used to detect intrusive network traffic as they penetrate a digital computer network \cite{javaid2016deep}. They aim to preserve the three key principles of information security; confidentiality, integrity, and availability \cite{whitman2011principles}. NIDS scan and analyse the incoming traffic for any malicious indicators that may present a threat or harm to the target network. There are two main types of NIDS; 1) signature-based NIDS, which operates by scanning for a set of previously known attack rules or Indicators Of Compromise (IOC) \cite{ashoor2011importance} such as source/destination IPs and ports, hash values or domain names in an incoming network feed. This traditional method works efficiently against known attack scenarios where the complete set of IOC has been previously identified and registered within the NIDS. However, signature-based NIDSs have been vulnerable to zero-day attacks where there is a lack of knowledge of the IOCs related to the occurrence of the activity \cite{garcia2009anomaly}. In addition, the detection of modern advanced and persistent threats such as Cobalt Strike \cite{vandetecting} requires a sophisticated depth of behavioural change monitoring \cite{bhatt2014towards}, where the usage of traditional IOC is not sufficient in their detection. Therefore, the focus of NIDS development has shifted towards the modern type of NIDS with enhanced Machine Learning (ML) capabilities \cite{sarhan2021feature}.

ML is an emerging branch of Artificial Intelligence (AI) extensively used with great success to empower decision making systems across various domains \cite{goodfellow2016machine}. ML models operate by extracting and learning meaningful patterns from historical data during the design process. The models then apply the learnt semantics to classify or predict unseen data samples into their respective classes or values. The intelligence capability of ML has motivated its usage in many industries to provide a deeper level of analysis to automate and assist in complex decision-making tasks \cite{jordan2015machine}. ML enhances the performance and efficiency of hosting systems without being explicitly programmed \cite{mahesh2020machine} by learning complex patterns that are not trivial to recognize by domain experts. As such, ML has been welcomed in the development of NIDS to overcome the limitations faced by signature-based NIDS and to improve the detection of cyberattacks using an intelligent layer of defence \cite{tsai2009intrusion}. ML-based NIDS capabilities have been widely adopted in the security of modern computer networks to detect zero-day and advanced cyber threats. ML models are capable of learning the distinguishing semantic patterns between intrusive and benign network traffic and using it to detect any incoming traffic with malicious intent. Therefore, the focus on the network attacks' behavioural patterns and the lack of dependency of identified IOC to detect attacks \cite{bhuyan2013network} has attracted attention towards the development of ML-based NIDS.

The development phase of ML-based NIDS generally involves two key stages \cite{goodfellow2016machine}; the learning stage where the designed ML model learns the semantic patterns from intrusive and benign traffic, followed by a testing stage where the model classifies unseen data into either unsafe or safe classes. Standard ML techniques usually follow either a localised or a centralised learning method \cite{briggs2021federated}. A localised learning method involves local data samples collected from a single endpoint, the learning and testing occur locally on the endpoint \cite{youssef2020localised}, where it is generally more effective with a larger amount of data. This method often provides a high detection accuracy over Independent and Identically Distributed (IID) \cite{clauset2011brief} data samples with a similar probability distribution to the local data samples. However, as network traffic is often heterogeneous in nature \cite{kato2016deep} due to a multitude of safe applications/services and malicious threats/intrusions, localised training approaches do not scale well with the rapidly increasing and changing world traffic \cite{bhole2005measurement}. This is mainly due to the fact that the learning model is exposed to a limited variety of network traffic scenarios, hence it has a limited experience of other instances. As a result, modern research works have adopted centralised learning methods to overcome some of the limitations faced by localised approaches.

Centralised learning is widely proposed in the academic field where local data samples are collected from various endpoints and transmitted to a central server \cite{nardi2021centralised}. The central entity holds all data samples ideally representing an overall representation of the organisation's Standard Operating Environment (SOE) \cite{aupek2006architectural} and a large number of attack scenarios. The learning and testing stages are conducted on the central server where the learning models experience and extract useful patterns from heterogeneous network traffic. Therefore, enabling NIDSs to effectively detect network intrusions across non-IID data samples \cite{abbasi2021deep}. This method generally achieves a reliable detection performance across the various endpoints of an organisation. However, centralised learning requires the direct sharing of data samples between endpoints and a central entity \cite{yang2019federated}. This presents serious privacy and security concerns due to the nature of the data transmitted. Network data often contain sensitive information related to users' browsing sessions, applications and services utilised, often revealing critical endpoint details. As a result, a new and active research area known as federated learning \cite{li2020federated} has been utilised to overcome the privacy concerns of centralised learning methods.

Federated learning is an advanced technique of ML designed to address certain limitations of traditional ML. A federated learning setup allows for training a learning model across multiple decentralised endpoints, holding local data samples without exchanging them \cite{yang2019federated}. The key benefit of following a federated learning approach is preserving and maintaining the privacy and security of local data samples as they are no longer shared with other entities \cite{truex2019hybrid}. In addition, due to a lack of a central entity storing all data samples, there is lower latency, power and storage requirements due to the reduced transmission of data \cite{yang2020federated}. This is often a motivation for usage in IoT networks where federated learning has been widely adopted \cite{imteaj2021survey}. In the context of NIDS, this enables the design of smarter ML models as they are exposed to a large number of heterogeneous data samples generated using various endpoints while ensuring the privacy of network users \cite{preuveneers2018chained}. Federated learning approaches generally begins a global sever initiating an ML model. Each participating local client downloads a copy of the global model locally and improves it by conducting a learning process using the local data samples. Finally, the set of trained model's parameters is uploaded back to the global server where it is aggregated with each set of received parameters to improve the global model \cite{hard2018federated}, this presents a single federated learning round and can be repeated several times to reach a desirable performance.

In this paper, we take the application of machine learning a step further, where each local client is observed as a single organisation with a unique network of heterogeneous endpoints. While this attracts a great number of benefits, it also raises certain challenges which we address in this paper. The proposed federated learning methodology allows multiple organisations to share Cyber Threat Intelligence (CTI) \cite{brown2019evolution} to collaboratively and securely design an effective ML-based NIDS. This increases the exposure of the NIDS model to a multitude of network traffic scenarios including various benign SOEs and malicious attack scenarios occurring in different organisational networks. This is an important aspect of a real-world implementation as it causes each network to have its unique statistical distribution \cite{layeghy2021benchmarking} that ML models performance might not generalise across. Unlike centralised learning approaches, federated learning enables the collaboration between organisations to design shared ML-based NIDSs while keeping the privacy of their local data preserved. Decoupling the ability to learn from other organisations' network intelligence and attack experiences from the need to directly share sensitive data. Similarly to the traditional federated learning methodology, a single global organisation is required to orchestrate the whole process by initiating a global ML model. Each participating organisation downloads a copy of the global model locally, and trains it using its local data samples internally and uploads the updated parameters back to the global organisation. The global organisation receives the trained parameters from each participant and aggregates them to improve the global model before sending it back to each organisation for usage. This presents a single federared learning round and can be repeated across all participants to reach a reliable state of performance.

The outcome of the proposed method is a common and robust ML-based NIDS not limited to per a single organisation experience and available local training samples. The enhanced model is trained on heterogeneous data collected over a variety of non-IID networks \cite{zhao2018federated}, each presenting its unique behaviour of benign and a wide range of different attacks. In addition, the training data remains secure and preserved internally within each organisation perimeters without the explicit exchange of data samples. The key contributions of this paper are the proposal of a novel privacy-preserving cyber threat intelligence scheme and the evaluation of its performance using two key NIDS datasets. The results are analysed and compared to centralised and localised use cases to demonstrate the effectiveness of the scheme. In Section \ref{rw}, we explore some of the key related works and highlight their limitations, the motivation and benefits of the proposed intelligence sharing scheme are discussed in Section \ref{main}. In Section \ref{ev}, we perform an empirical evaluation and comparison of a collaboratively designed ML-based NIDS to demonstrate the robustness and benefits of the proposed framework. Finally, we conclude this paper in Section \ref{con} and list some of the critical future works.

\section{Related Works}
\label{rw}
A large number of research papers have aimed to adopt a federated learning approach in the design of ML-based NIDS. While most papers, focused on the structure and parameters of the adopted learning model, all of the training and evaluation stages were conducted using a single organisational network data split over several local endpoints. Therefore, the data samples used in the learning model are not very different in nature as they all originate within the same network. To the best of our knowledge, no paper has considered the requirements of designing an ML-based NIDS using several heterogeneous data sources collected across multiple non-IID networks. This framework is to be ideally used to provide a collaborative CTI mechanism to organisations without the direct sharing of sensitive data. The outcome would be a model capable of distinguishing between several benign traffic and malicious attack behaviours. A centralised ML technique requires the exchange of data between organisations which is very challenging due to the security and privacy aspects. Therefore, the collaborative scheme requires to be conducted in a federated learning manner to achieve the same outcome without the need of sharing data.

In \cite{rahman2020internet}, Abdul Rahman et al. evaluated the detection performance of a NIDS designed in a centralised, on-device, and federated learning use case. In a centralised scenario a single entity stores and performs analysis on data collected across multiple devices. The on-device use case is a self-learning technique where a single device analyses its local data samples. The comparison was conducted using safe and malicious network data samples in the NSL-KDD dataset. The results show that federated learning outperforms the on-device learning method and achieves a similar detection performance in a centralised manner while maintaining the privacy of local data samples.

Mothukuri et al.\cite{mothukuri2021federated} explored different parameters of a federated learning-based anomaly detection approach to detect IoT intrusions using decentralised on-device data. The paper explored two deep learning models; Long Short Term Memory (LSTM) and Gated Recurrent Units (GRU) with various window sizes and an additional Random Forest ensemble component to combine the predictions from different layers. The evaluation was conducted on the Modbus-based dataset which consists of benign IoT telemetry traffic and four attack scenarios. The results show that their approach outperformed the centralised ML approach with a minimised error detection rate and reduced the number of false alarms.

In paper \cite{popoola2021federated}, Popoola et al. proposed a deep neural network model to detect zero-day botnet traffic with a high classification performance. By following a federated learning approach, the method guarantees to preserve data privacy and security, in addition, it has a lower communication overhead, network latency and memory space for storage of training data. The paper explored sixteen DNN models to determine the optimal neural architecture for efficient classification. The traditional FedAvg algorithm \cite{mcmahan2017communication} is used for the aggregation of local model parameters. The performance of the federated learning methodology in the detection of zero-day botnet attacks is compared with centralised and localised methods where the federated learning achieves similar performance to the centralised method while preserving data privacy. 

Overall, recent research works have addressed different aspects of the federated learning process, which is an active research area, such as communication cost, privacy, security, and resource allocation. However, each endpoint utilised is holding a set of local data samples that represent a similar distribution to the overall data. These approaches may not scale well with the rapid growth of network attacks that may compromise organisational networks. In the real world, the local network data samples in each organisation are unique in their statistical distribution depending on the organisations SOE and the types of malicious threats faced. Therefore, different datasets with non-IID samples are required to simulate a federated learning approach across multiple organisations. Therefore, we aim in this work to investigate the applicability of collaborative CTI sharing based on federated learning for network intrusion detection.

\section{Cyber Threat Intelligence Sharing}
\label{main}

Data is considered the most valuable and powerful tool that an organisation might own in the 21st century. A lot of organisations in many sectors depend on data to provide insights and extract meaningful patterns through analytical engines. ML has provided organisations with intelligent algorithms able of extracting and learning semantic attributes from historical data \cite{mahesh2020machine} to provide insights into the prediction or classification of future data. As such, ML capabilities have been adopted in the design of NIDSs to monitor and preserve the digital perimeters of organisations' networks. To achieve this goal, network data traffic has been captured from organisational networks to design an ML model. During the training process, the model learns the distinguishing patterns between benign and intrusive traffic which can be used in future detection. ML-based NIDS has been proven to be reliable in the detection of zero-day and modern attacks by utilising the malicious behaviour and attack chains rather than a set of IOCs implemented in signature-based NIDS.

\subsection{Motivation}
A large number of research work has been conducted to improve the overall performance of ML-based NIDS. Current traditional systems have generally been designed in a localised ML manner where models learn traffic patterns from a single network source. This method provides the learning model with high visibility into a target organisational network's SOE activities, and malicious threats encountered in the past. However, as an ML model only knows what it learns, traditional ML-based NIDS are limited to per an organisation experience independently and might be incapable to generalise across non-IID network sources. There is a high chance of varying distributions in different networks due to the unique SOEs and their associated threats implemented within organisations. This presents a significant risk to organisations due to the rapidly changing network environments caused by modern work practices such as new services or an incoming advanced threat such as zero-day attacks.

Therefore, the current method of ML-based NIDS design does not scale with the rapid growth of network benign and attack variants as there is a requirement to collect each scenario in the target organisation. We utilise the change of networks as a baseline in our experiments, i.e., when an ML model is trained on one network source and evaluated on a different network. This measures how well a learning model generalises across other networks. Another key limitation of current approaches is the requirement to collect a large amount of training data samples to increase the performance and generalisation of the ML model and avoid overfitting over a few data samples \cite{dietterich1995overfitting}. Therefore, particularly in the design of ML-based NIDS, following a supervised method adopted in this paper, a large number of labelled benign and attack data samples are required. The lack of labelled training data is a major challenge for small organisations aiming to effectively design an intrusion detection model.

Due to the lack of shared intelligence, organisations can not benefit from the patterns of safe application or malicious intrusions occurring in other organisations. Therefore, a collaborative ML approach between organisations is necessary for the design of enhanced NIDS. The three ML scenarios are considered for this purpose where the traditional endpoints are replaced with organisations. The localised learning method is inapplicable as it involves a single source of organisational data. This is used for comparison purposes in this paper as a non-collaborative scenario where an organisation does not share intelligence. The centralised learning scenario requires a direct sharing of data between organisations and a central entity to allow for the training of an ML model. This method enables the learning model to extract useful patterns from various data samples collected over the participating organisational networks to overcome the issues faced in the localised learning scenario.

However, network data often present sensitive information such as user browsing sessions, applications accessed and critical endpoint details, e.g. domain controllers. Therefore, following a centralised learning approach poses privacy, security, and transactional risks that organisations would generally avoid. Moreover, recent strict laws such as the General Data Protection Regulation (GDPR) \cite{zarsky2016incompatible} are enforced to protect consumers data privacy and address concerns related to unauthorised sharing of user-related information. The violation of privacy conserving regulations often presents serious legal concerns and hefty fines of up to \$20million \cite{8190804} in the case of a GDPR breach. Unfortunately, centralised learning requires a central entity to collect, store and analyse network data samples collected from participating organisations, which could make it unfeasible to conduct in the real world.

It is important to note that the sharing of CTI is not uncommon in the security field. In fact, many organisations using signature-based NIDS heavily rely on CTI platforms, such as Malware Information Sharing Platform (MISP) \cite{wagner2016misp} a widely-used open-source platform. CTI platforms develop utilities and documentation for more effective threat intelligence by the sharing of IOCs related to external threat actors. Organisations generally integrate an intelligence feed with their traditional signature-based NIDS to provide a high detection accuracy against the associated attacks. However, in ML-based NIDS, the sharing of attack data samples might include revealing information related to the targeted user, endpoint or application depending on the attributes provided.

\subsection{Collaborative Federated Learning}

\begin{figure*}[ht]
\begin{subfigure}{.2\textwidth}
  \centering
  \includegraphics[width=3cm, height=2cm]{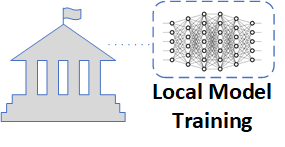} 
  \caption{Localised}
  \label{ll}
\end{subfigure}
\hfill
\begin{subfigure}{.3\textwidth}
  \centering
  \includegraphics[width=5cm, height=4.5cm]{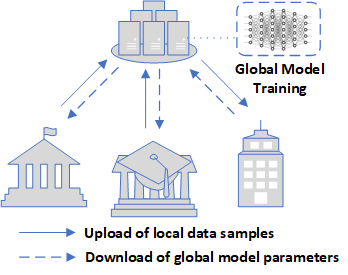}
  \caption{Centralised}
  \label{cc}
\end{subfigure}
\hfill
\begin{subfigure}{.4\textwidth}
  \centering
  \includegraphics[width=7cm, height=4.5cm]{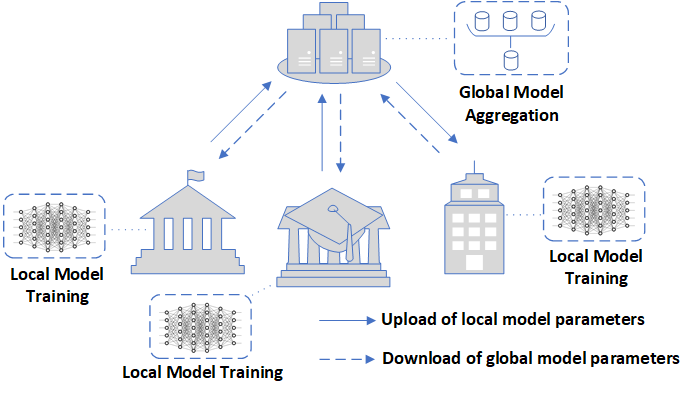}
  \caption{Federated}
  \label{ff}
\end{subfigure}
\caption{Machine Learning Scenarios}
\label{mm}
\end{figure*}

To overcome the previously mentioned limitations, CTI sharing between organisations via a federated learning approach is required to increase the knowledge base of the learning models while maintaining the privacy of user information. The three learning scenarios are illustrated in Figure \ref{mm}. The learning model is exposed to a wider range of benign and attack variants in order to achieve reliable detection accuracy across previously unseen traffic in a given organisation. The proposed framework allows organisations to join forces by sharing their cyber intelligence and insights. In addition, organisations that do not collect and store a sufficient amount of network traffic required for the training of a learning model are now able to design an effective ML-based by collaborating with other organisations. As each participant contributing with a minimum amount of data samples would permit the design of a successful system, our approach tackles the data scarcity problem and makes it possible to design an ML-based NIDS without the need to collect a large amount of training data.

Moreover, by adopting a federated learning approach, the local network data samples remain distributed internally across the organisations, hence persevering the privacy and integrity of sensitive users' network information. A federated learning setup includes a global server that coordinates and orchestrates the independent training of the local models. In this paper the global server is hosted within a participant organisation, however, this framework enables it to be hosted externally within a trusted mediator or a blockchain, which is part of future works. One of the main requirements of this framework is for each participating organisation to hold its local network data traffic in a common logging format. The benefits of having a standard feature set are many and explained here \cite{sarhan2021explainable} and \cite{portmann2021netflow}. In this framework, a common feature set enables streamlined federated learning as the global model can extract meaningful patterns across a standard set of data features. The global model is designed to be compatible with the agreed network logging format.

Similar to standard federated learning approaches, the process is triggered by a global server initiating an ML model with a pre-defined architecture and parameters. The model is forwarded to each participant before it is trained and enhanced locally using the internal network data samples. Only the updated weights are sent back to the global server. The global server follows the \textit{FedAvg} technique \cite{mcmahan2017communication} defined in Algorithm \ref{alg1}. In this step, the server aggregates the weights uploaded by each organisation to generate an enhanced intrusion detection model with an improved set of parameters designed over each participant's network. This presents a single federated learning round and can be repeated several times to achieve a better detection performance across all the environments.

\begin{algorithm}[!h]\small
  \SetKwData{Left}{left}\SetKwData{This}{this}\SetKwData{Up}{up}
  \SetKwFunction{Union}{Union}\SetKwFunction{FindCompress}{FindCompress}
  \SetKwInOut{Global}{Global Server Executes}\SetKwInOut{Local}{Local Organisation Update $(k,w)$}

\Indp\Indpp

  \Global{   $  \text{Initialize } w_0$\\
  \For{each federated learning round t = 1,2,...:}{
    $S_{t} \leftarrow(\text {Set of } K \text { organisations) } $
    
     \For {each organisation $k \in S_{t}$ :}{
$w_{t+1}^{k} \leftarrow \text { Local Organisation Update } (k, w_{t}) $

 $w_{t+1} \leftarrow \sum_{k=1}^{K} \frac{m_{k}}{m} w_{t+1}^{k}$}}
  }
  \Local{
Run on organisation $k$ \\
$\mathcal{B} \leftarrow\left(\right.$ split $\mathcal{P}_{k}$ into batches of size $B$ ) \\
\For{each local epoch $i$ from 1 to $E$:}{
\For{batch $b \in \mathcal{B}$:}{
$w \leftarrow w- $n$ \nabla $l$ (w ; b)$ 

return $w$ to server}}
    }

\caption{FederatedAveraging. The $K$ organisations are indexed by $k$; $B$ is the local training batch size, $E$ is the number of local epochs, $\mathcal{P}$ is the local training set, $l$ is the loss of prediction on example $(x_i,y_i)$, $n$ is the local learning rate, and $m$ is the global learning rate. }
\label{alg1}
\Indp\Indpp

\end{algorithm}\DecMargin{1em}

The key outcome is the design of a robust ML-based NIDS obtained from a collaboration between organisations without the need to share information with other participants to preserve data privacy. The final model is capable of detecting a wider range of attacks originating from several sources which are crucial in an organisational defence system. This provides a robust learning model with global intelligence and insights able to distinguish between heterogeneous benign and attack traffic. Such smart models would possibly lead to a lower false alarm rate in case of a variation of the benign traffic distribution caused by a modification of the SOE due to the learning from several networks' network usage. Moreover, a higher detection rate of advanced and zero-day attacks is promising due to the extraction of malicious patterns from a wider range of attacks that targeted several organisational networks.


\section{Experimental Setup}
\label{ev}
To evaluate the feasibility and performance of our proposed collaborative CTI Sharing scheme based on federated learning for NIDS, we utilise two key and widely used NIDS datasets. Each dataset has been collected over a different network, each consisting of a different set of benign applications and malicious attack scenarios. Therefore, each dataset represents a certain organisational network with a unique SOE and malicious events encountered. The datasets also hold a very distinctive statistical distribution as presented here \cite{layeghy2021benchmarking}. This matches the assumption of obtaining non-IID datasets collected over different real-world networks. While the datasets are unique in their applications, protocols, and data classes, they share a common feature set based on NetFlow v9 \cite{claise2004cisco}, a defacto standard protocol in the networking industry. In this paper, the NF-UNSW-NB15-v2 and NF-BoT-IoT-v2 datasets are used to simulate two organisations collaborating in the design of a universal ML-based NIDS. By following a federated learning-based technique, each dataset is maintained internally in the learning and testing stages. The key benefit of the federated learning collaboration is the design of a robust ML-based NIDS trained on multiple organisational networks without any exchange of local network data samples. Hence, maintaining the privacy and integrity of sensitive user's information. The datasets structure and format are explained below;

\begin{itemize}
    \item \textbf{NF-UNSW-NB15-v2} \cite{sarhan2021towards}- A NetFlow-based dataset released in 2021 by the University of Queensland that contains nine attack scenarios; Exploits, Fuzzers, Generic, Reconnaissance, DoS, Analysis, Backdoor, Shellcode, and Worms.  The dataset is generated by converting the publicly available pcap files of the UNSW-NB15 dataset \cite{moustafa2015unsw} to 43 NetFlow v9 features using the nprobe \cite{deri2003nprobe} tool. The complete number of data flows are 2,390,275 out of which 95,053 (3.98\%) are attack samples and 2,295,222 (96.02\%) are benign. The source dataset (UNSW-NB15) is a widely used NIDS dataset in the research community. UNSW-NB15 was released in 2015 by the Cyber Lab of the Australian Center for Cyber Security (ACCS). The IXIA Perfect Storm tool was configured to simulate benign network traffic and synthetic attack scenarios.  
    
    \item \textbf{NF-BoT-IoT-v2} \cite{sarhan2021towards}- An IoT NetFlow-based dataset released in 2021 by the University of Queensland that contains four attack scenarios; DDoS, DoS, Reconnaissance, and Theft. The dataset is generated by converting the publicly available pcap files of the BoT-IoT \cite{koroniotis2019towards} dataset to 43 NetFlow v9 features using the nprobe \cite{deri2003nprobe} tool. The complete number of data flows are 37,763,497 labelled network data flows, where the majority are attack samples; 37,628,460 (99.64\%) and, 135,037 (0.36\%) are benign. The source dataset (BoT-IoT) is generated by an IoT-based network environment that consists of normal and botnet traffic. BoT-IoT was released in 2018 by the Cyber Range Lab of the ACCS. The non-IoT and IoT traffic were generated using the Ostinato and Node-red tools, respectively, and Tshark is used to capture the network packets.

\end{itemize}

\subsection{Evaluation}

Three different approaches are considered in the evaluation process; federated, centralised, and localised learning scenarios. In the federated learning use case, an agreement is made between two or more participant organisations. The agreement specifies that each organisation participates by downloading an initialised ML model locally from a global server. The participant would train the learning model on its local data samples internally and upload the trained model parameters back to the global server. The global server receives each parameter from each organisation and averages the parameters together into a single model. This represents a single federated learning round and can be repeated until the model performance reaches a reliable state. For the centralised learning scenario, a similar collaboration agreement is conducted, however, each participating organisation sends their local training data to a central server for the training of the ML model on the complete set of aggregated data. In the localised learning scenario, there are no collaborations between organisations, therefore the model is trained on each organisation's limited local data samples.

\begin{table}[h]\footnotesize	
\centering
\caption{Evaluation Metrics}
\begin{tabular}{|>{\centering\arraybackslash}m{4cm}|>{\centering\arraybackslash}m{5.5cm} |>{\centering\arraybackslash}m{3cm} |}

\hline
\textbf{Metric}            & \textbf{Definition}           & \textbf{Equation}                               \\ \hline
Accuracy                   & The percentage of correctly classified samples.              &\normalsize $\frac{TP+TN}{TP+FP+TN+FN} \times 100$\\ \hline
Detection Rate (DR)        & The percentage of correctly classified total attack samples. &\normalsize $\frac{TP}{TP+FN} \times 100 $\\ \hline

False Alarm Rate (FAR)     & The percentage of incorrectly classified benign samples. &  \normalsize $\frac{FP}{FP+TN} \times 100$   \\ \hline

Area Under the Curve (AUC) & The area underneath the DR and FAR plot curve.        & N/A       \\ \hline
F1 Score                   & The harmonic mean of the model's precision and DR.  & \normalsize $2 \times \frac{DR\;\times \;Precision}{DR\; +\; Precision}$         \\  \hline 
Time                & The time required in seconds to complete the training of the ML model.  &  N/A         \\  \hline 
\end{tabular}%
\label{evm}
\end{table}
\renewcommand{\arraystretch}{1}

The evaluation metrics used to evaluate the performance of the ML models are defined in Table \ref{evm}. The metrics are calculated in a binary format based on the True Positive (TP) and True Negative (TN) representing the numbers of the correctly classified attack and benign data samples, respectively. In addition to the False Positive (FP) and False Negative (FN) represent the numbers of the incorrectly classified benign and attack data samples, respectively. The experiments were conducted using Google's Tensorflow Federated (TFF) framework \cite{tensorflow_federated_2021} for the federated learning scenario and Tensorflow framework \cite{tensorflow_2021} for the centralised and localised scenarios. The datasets are reprocessed by dropping the flow identifiers such as source/destination IP and ports to avoid bias towards the attacking and victim end nodes. A subset of each dataset with an equal number of benign and attack samples has been used to generate the below results. Each dataset has been split into training and testing splits in a ratio of 70\% to 30\%, respectively. A Min-Max scalier has been applied to normalise each dataset values into the same scale, defined as, 
\begin{equation}X_*=\frac{X-X_{min}}{X_{max}-X_{min}}\end{equation}
\noindent where $X\textsubscript{*}$ is the output value ranging from 0 to 1, $X$ is the input value and $X\textsubscript{max}$ and $X\textsubscript{min}$ are the maximum and minimum values of the feature respectively. The parameters used in this paper to design the ML experiments are represented in Table \ref{par}.

\begin{table}[h]\footnotesize
\centering
\begin{tabular}{|l|l|}
\hline
\multicolumn{1}{|c|}{\textbf{Parameter}} & \multicolumn{1}{c|}{\textbf{Value}} \\ \hline
Local Epochs                             & 3                                   \\ \hline
Batch Size                               & 2048                                \\ \hline
Local Optimiser                          & Adam                                \\ \hline
Local Learning Rate                      & 0.001                               \\ \hline
Loss Function                            & Binary Crossentropy                 \\ \hline
Federated Learning Rounds*                                & 10                                 \\ \hline
Server Optimiser*                         & Adam                                \\ \hline
Server Learning Rate*                     & 0.05                                \\ \hline

\end{tabular}
\caption{Experimental Parameters. *Only applies to federated learning}
\label{par}
\end{table}

It is important to note that while a discovery stage was conducted by exploring a large number of hyper-parameter sets to obtain a reliable detection performance, the full exploration of the parameter's space is not covered in this paper. The performance of the ML models and the overall proposed scheme can be improved by optimising the set of parameters adopted. Two key ML models adopted in the ML-based NIDS have been designed to demonstrate the effectiveness of the proposed framework. The same parameters were used across the three scenarios for a fair comparison. A Deep Neural Network (DNN) and Long Short-Term  Memory (LTSM) have been used in this paper with their parameters defined in Table \ref{par2}. The hyper-parameters were identically designed to provide a fair comparison of their performance. There is a dropout of 40\% of the input units between each hidden layer to help prevent over-fitting of the local client's data.

\begin{table}[h]\footnotesize
\centering
\begin{tabular}{l|l|l|}
\cline{2-3}
\multicolumn{1}{c|}{\textbf{}}                & \multicolumn{1}{c|}{\textbf{Nodes}} & \textbf{Activation Function} \\ \hline
\multicolumn{1}{|l|}{\textbf{Input Layer}}    & 39  (Number of input features)                                & N/A                          \\ \hline
\multicolumn{1}{|l|}{\textbf{Hidden Layer 1}} & 12                                  & Relu                         \\ \hline
\multicolumn{1}{|l|}{\textbf{Hidden Layer 2}} & 6                                   & Relu                         \\ \hline
\multicolumn{1}{|l|}{\textbf{Hidden Layer 3}} & 3                                   & Relu                         \\ \hline
\multicolumn{1}{|l|}{\textbf{Output Layer}}   & 1                                   & Sigmoid                      \\ \hline
\end{tabular}
\caption{Model Parameters}
\label{par2}
\end{table}

\subsection{Results}
\begin{figure*}[h]
\begin{subfigure}{.5\textwidth}
  \centering
  \includegraphics[width=7cm, height=4.5cm]{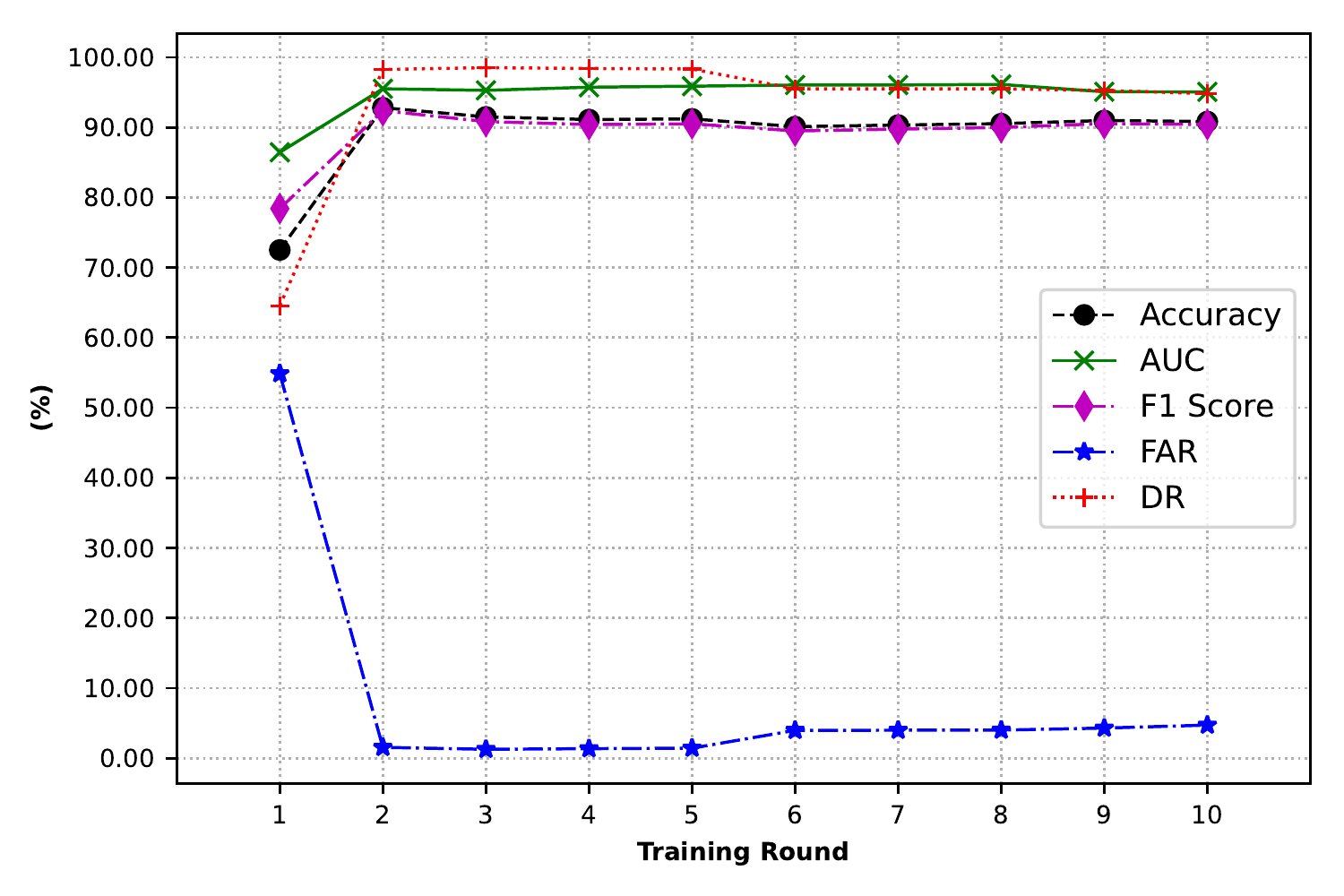} 
  \caption{NF-UNSW-NB15-v2}
  \label{u}
\end{subfigure}
\hfill
\begin{subfigure}{.5\textwidth}
  \centering
  \includegraphics[width=7cm, height=4.5cm]{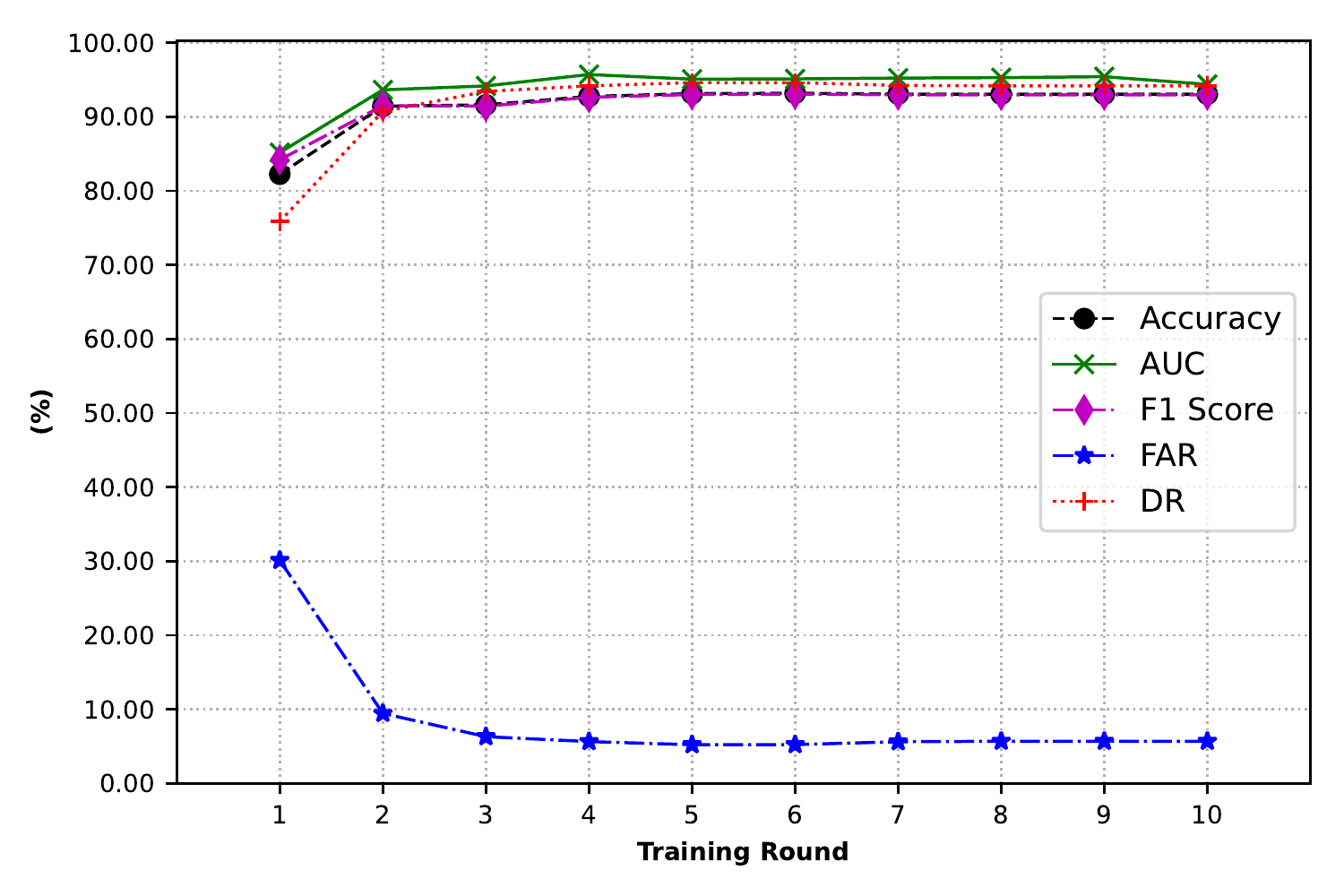}
  \caption{NF-BoT-IoT-v2}
  \label{uu}
\end{subfigure}
\caption{Federated Learning using a DNN Model}
\label{tt}
\end{figure*}

\begin{figure*}[h]
\begin{subfigure}{.5\textwidth}
  \centering
  \includegraphics[width=7cm, height=4.5cm]{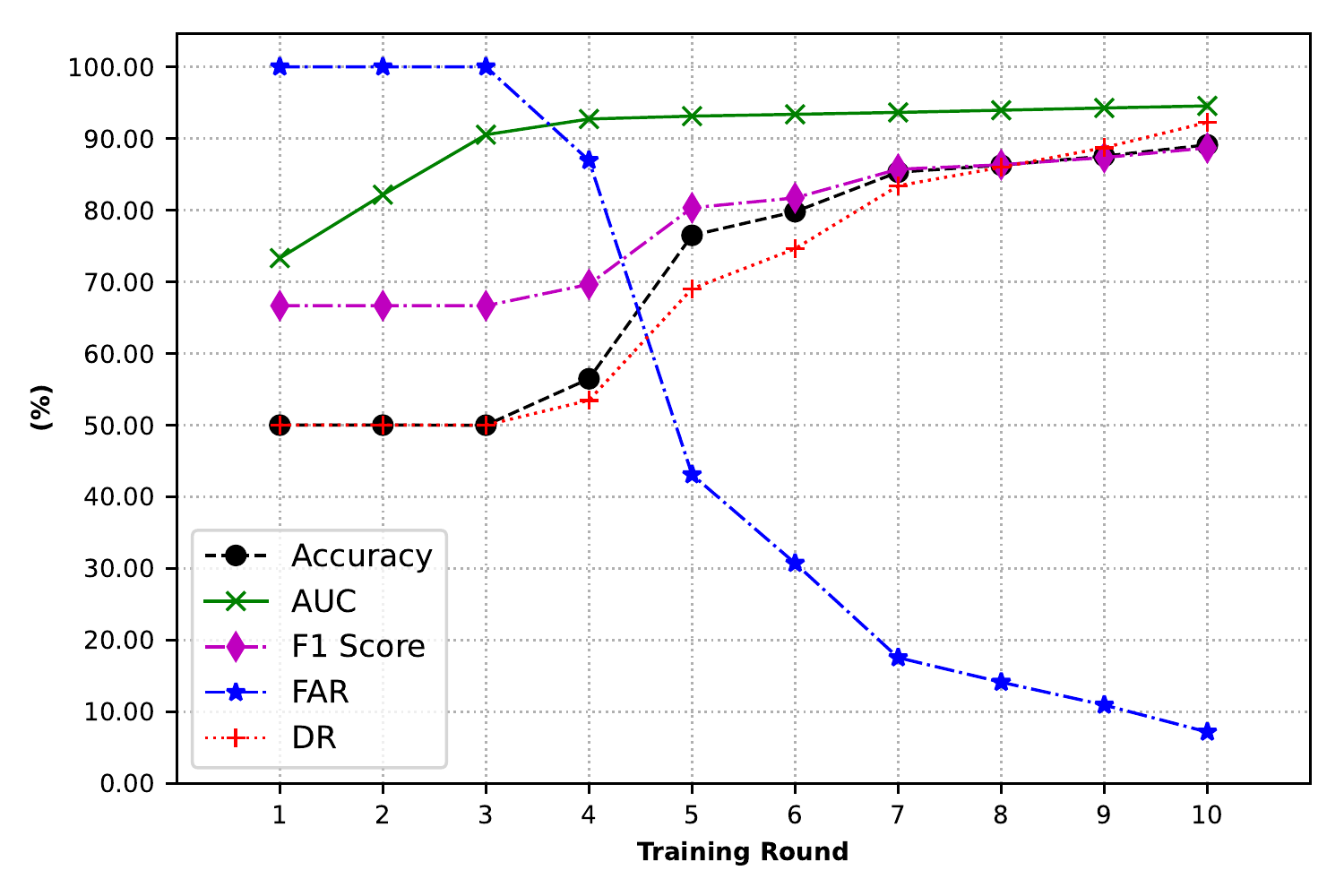} 
  \caption{NF-UNSW-NB15-v2}
  \label{i}
\end{subfigure}
\hfill
\begin{subfigure}{.5\textwidth}
  \centering
  \includegraphics[width=7cm, height=4.5cm]{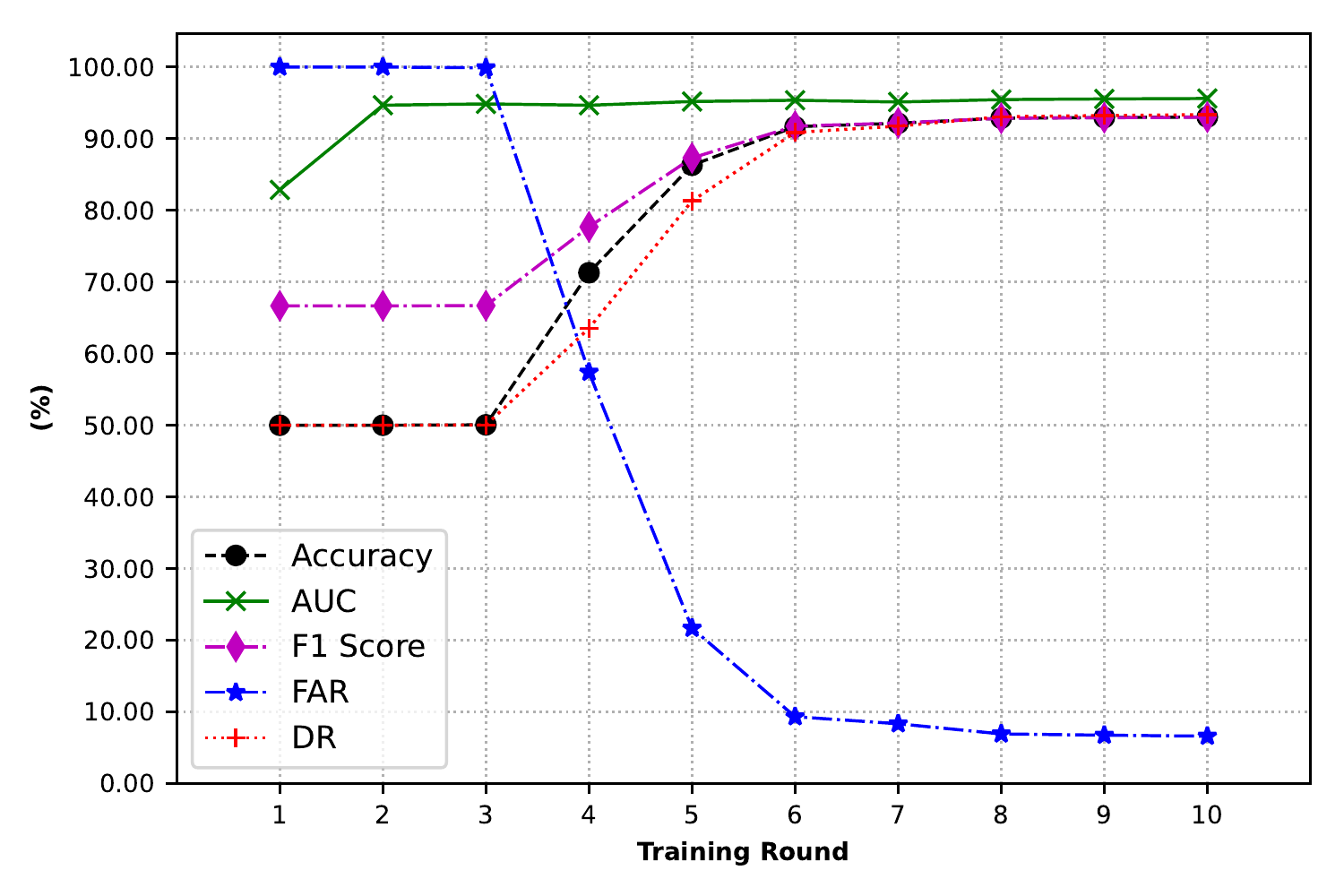}
  \caption{NF-BoT-IoT-v2}
  \label{ii}
\end{subfigure}
\caption{Federated Learning using an LSTM Model}
\label{yy}
\end{figure*}

The results in this section are collected over the test sets after the training has been conducted using the mentioned training scenario. We start with federated learning separately in Figures \ref{tt} and \ref{yy}, where the detection performance of the DNN and LSTM models respectively is evaluated on each dataset separately. The caption of each sub-figure identifies the test dataset used in the evaluation process. A set of results is collected after each federated learning round to analyse the improvement of the ML-based NIDS after each aggregation process. The results are plotted in line graphs where the percentage value is presented on the y-axis and the federated learning round number is listed on the x-axis, each line presents a different evaluation metric.

In Figure \ref{tt}, the DNN model achieves a reliable performance across the two datasets, where it rapidly converges to its maximum performance after the third round and it stabilises thereafter. There is a significant drop in the FAR in both datasets after the first federated learning round in both datasets. The remaining metrics increase by around 30\% and 15\% in the NF-UNSW-NB15-v2 and NF-BoT-IoT-v2 datasets respectively. In Figure \ref{yy}, the LSTM model required a larger number of federated learning rounds to reach a reliable detection performance. During the first three rounds, the model was achieving a poor performance of 50\% accuracy in both datasets. However, the performance rapidly increased between the fourth and seventh-round until it converged to its maximum reliable performance. The FAR dropped from 100\% to nearly 8\% during the 10 rounds of federated learning in both datasets.

Tables \ref{unsw} and \ref{bot} compare the three training scenarios by displaying the full set of evaluation metrics achieved on the NF-UNSW-NB15-v2 and NF-BoT-IoT-v2 test datasets, respectively. The results are grouped by the ML used and the scenario followed in the training process. In addition, the time required to complete the training stage is measured in seconds. In the federated learning scenario, the results achieved after the tenth round are presented in the tables. It is important to note that for the federated learning scenario, the time is measured over ten rounds, which might not be required to achieve a reliable performance as demonstrated in Figure \ref{tt}.

\begin{table}[h]\scriptsize
\centering
\begin{tabular}{ll|l|l|l|l|l|l|}
\cline{3-8}
                                                     &                      & \textbf{ACC} & \textbf{AUC} & \textbf{F1} & \textbf{DR} & \textbf{FAR} & \textbf{Time (s)} \\ \hline
\multicolumn{1}{|l|}{\multirow{3}{*}{\textbf{DNN}}}  & \textbf{Federated}          &   90.83           &       95.06       &   90.40          &       94.82      &     4.72         &     34.21          \\ \cline{2-8} 
\multicolumn{1}{|l|}{}                               & \textbf{Centralised} &    99.38          &     99.47         &    99.38         &     99.42        &  0.67            &    5.83           \\ \cline{2-8} 
\multicolumn{1}{|l|}{}                               & \textbf{Localised}    &   51.34           &  59.89            &   7.89          &     4.17        &      1.48        &     3.77          \\ \hline
\multicolumn{1}{|l|}{\multirow{3}{*}{\textbf{LSTM}}} & \textbf{Federated}          &  89.10            &    94.55          &     88.69        &       92.25      &     7.17         &    51.92           \\ \cline{2-8} 
\multicolumn{1}{|l|}{}                               & \textbf{Centralised} &       95.80       &      98.46        &     95.65        &      92.55       &           0.96   &      9.57         \\ \cline{2-8} 
\multicolumn{1}{|l|}{}                               & \textbf{Localised}    &   52.32           &    79.75          &    10.82         &     5.78        &     1.15         &    7.19           \\ \hline
\end{tabular}
\caption{NF-UNSW-NB15-v2: Binary-class Detection}
\label{unsw}
\end{table}

In Table \ref{unsw}, the binary-class detection results achieved on the NF-UNSW-NB15-v2 dataset are presented, where the federated and centralised learning scenarios achieve a reliable performance of 90.83\% and 99.38\% accuracy using the DNN model and 89.10\% and 95.80\% using the LSTM model, respectively. The lower performance noted in the federated learning approach was mainly due to a higher number of FAR of 4.72\% and 7.17\% using the DNN and LSTM models compared to 0.67\% and 0.96\% in the centralised scenario. While in the localised learning scenario, the lowest training time was achieved due to a lower number of training samples by a single organisation, the attack detection performance is unreliable. The model was unable to detect most of the attacks present in the NF-UNSW-NB15-v2 dataset after being trained on the NF-BoT-IoT-v2 dataset achieving an inadequate DR of 4.17\% and 5.78\% using the DNN and LSTM models, respectively.

\begin{table}[h]\scriptsize
\centering
\begin{tabular}{ll|l|l|l|l|l|l|}
\cline{3-8}
                                                     &                      & \textbf{ACC} & \textbf{AUC} & \textbf{F1} & \textbf{DR} & \textbf{FAR} & \textbf{Time (s)} \\ \hline
\multicolumn{1}{|l|}{\multirow{3}{*}{\textbf{DNN}}}  & \textbf{Federated}          &    93.04          &   94.37           &   92.95          &      94.16       &    5.69          &     34.21          \\ \cline{2-8} 
\multicolumn{1}{|l|}{}                               & \textbf{Centralised} &     93.83         &      96.74        &    93.84         &     93.99        &    6.32          &      5.83         \\ \cline{2-8} 
\multicolumn{1}{|l|}{}                               & \textbf{Localised}    &   86.21           &  86.89            &     86.92        &    91.66         &      19.25        &    3.10           \\ \hline
\multicolumn{1}{|l|}{\multirow{3}{*}{\textbf{LSTM}}} & \textbf{Federated}          &    93.03          &       95.61       &       93.00      &    93.36         &     6.59         &      51.92         \\ \cline{2-8} 
\multicolumn{1}{|l|}{}                               & \textbf{Centralised} &     93.90         &       94.76       &      93.76       &     91.71        &   3.92           &       9.57        \\ \cline{2-8} 
\multicolumn{1}{|l|}{}                               & \textbf{Localised}    &  88.52            &   88.87           & 88.87            &    91.66         &       14.62       &   6.62            \\ \hline
\end{tabular}
\caption{NF-BoT-IoT-v2: Binary-class Detection}
\label{bot}
\end{table}

In Table \ref{bot}, the binary-class intrusion detection results collected over the NF-BoT-IoT-v2 test set are presented. A similar pattern in the NF-UNSW-NB15-v2 dataset is observed where the federated and centralised learning scenarios achieve a reliable intrusion detection performance. The accuracy achieved by the federated and centralised learning methods is 93.04\% and 9.83\% using DNN and 93.03\% and 93.90\% using LSTM, respectively. The attack DR is slightly higher using both ML models in the federated learning method compared to the centralised learning method. Surprisingly, the localised learning use case achieved significantly better results on the NF-BoT-IoT-v2 test set when trained on the NF-UNSW-NB15-v2 dataset. This was not the same case for the other way around. This could indicate the presence of meaningful patterns in NF-UNSW-NB15-v2 to help the model identify attacks in NF-BoT-IoT-v2. The accuracy achieved is 86.21\% using the DNN model and 88.52\% using the LSTM model, the performance drop is mainly caused by a high FAR of 19.25\% and 14.62\%, respectively.

\begin{table}[h]\scriptsize
\centering
\begin{tabular}{ll|l|l|l|l|l|l|l|l|l|l|}
\cline{3-12}
                                                     &                      & \textbf{Analysis} & \textbf{Backdoor} & \textbf{DoS} & \textbf{Exploits} & \textbf{Fuzzers} & \textbf{Generic} & \textbf{Recon} & \textbf{Shellcode} & \textbf{Worms} & \textbf{Average}\\ \hline
\multicolumn{1}{|l|}{\multirow{3}{*}{\textbf{DNN}}}  & \textbf{Federated}          &  83.62                 &    82.69               &       85.91       &        85.94           &       86.47           &   87.12               &      85.12                   &       88.39             &   82.35         &  85.06  \\ \cline{2-12} 
\multicolumn{1}{|l|}{}                               & \textbf{Centralised} &    100.00               &     99.23              &    98.22          &        99.15           &       99.51           &      99.84            &   99.82                      &     100.00               &     100.00      &   99.41  \\ \cline{2-12} 
\multicolumn{1}{|l|}{}                               & \textbf{Localised}    &   3.62                &   3.23                &     4.37         &       5.09            &      4.12            &     1.17             &   6.21                      &            2.10        &         6.12    &  4.00 \\ \hline
\multicolumn{1}{|l|}{\multirow{3}{*}{\textbf{LSTM}}} & \textbf{Federated}          &     83.62              &     82.99              &       84.78       &       85.09            &    85.69              &      85.78            &    84.28                     &    89.73                &       85.29   &   85.25   \\ \cline{2-12} 
\multicolumn{1}{|l|}{}                               & \textbf{Centralised} &      100.00             &     98.46              &     92.17         &        80.98           &      98.89            &      98.03            &     99.82                    &     100.00               &     100.00    &     96.48  \\ \cline{2-12} 
\multicolumn{1}{|l|}{}                               & \textbf{Localised}    &      4.35             &  5.22                 &   6.79           &     5.96              &       6.05           &        1.79          &  10.04                       &     3.74               &  16.33           &  6.70 \\ \hline
\end{tabular}
\caption{NF-UNSW-NB15-v2: Multi-class Detection}
\label{unsw2}
\end{table}

In Tables \ref{unsw2} and \ref{bot2}, we deep dive into the results of the NF-UNSW-NB-v2 and NF-BoT-IoT-v2 datasets to measure each attack DR separately in a multi-class manner. Similarly, the results are grouped by the ML used and the scenario followed in the training process, and the federated learning results are achieved after the tenth training round. In addition, we measure the average of the attack DR to compare the three scenarios based on the number of attack behaviours detected. In Table \ref{unsw2}, the highest DR is achieved by the centralised method in the NF-UNSW-NB15-v2 with an almost perfect DR of 99.41\% using the DNN model and 96.48\% using the LSTM model. The analysis, shellcode, and worm attacks were fully detected using both models. The federated learning approach came in second with an average DR of around 85\% using both models. As seen in previous results, the localised scenario is unreliable in the detection of any attacks in the NF-UNSW-NB15-v2 dataset with an average DR of 6.70\%.

\begin{table}[h]\scriptsize
\centering
\begin{tabular}{ll|l|l|l|l|l|}
\cline{3-7}
                                                     &                      & \textbf{DDoS} & \textbf{DoS} & \textbf{Recon} & \textbf{Theft} & \textbf{Average}\\ \hline
\multicolumn{1}{|l|}{\multirow{3}{*}{\textbf{DNN}}}  & \textbf{Federated}          &      91.66         &      92.24        &        91.96                 &   100.00        &   93.40  \\ \cline{2-7} 
\multicolumn{1}{|l|}{}                               & \textbf{Centralised} &    99.98           &     95.31         &       44.46                  &   100.00    &   84.94     \\ \cline{2-7} 
\multicolumn{1}{|l|}{}                               & \textbf{Localised}    &   98.04            &   92.97           &     36.33                    &      100.00   &   81.84    \\ \hline
\multicolumn{1}{|l|}{\multirow{3}{*}{\textbf{LSTM}}} & \textbf{Federated}          &     92.40          &      93.16        &         92.88                &     100.00   &  94.61     \\ \cline{2-7} 
\multicolumn{1}{|l|}{}                               & \textbf{Centralised} &    98.14           &     93.07         &         39.01                &   100.00     &  82.56      \\ \cline{2-7} 
\multicolumn{1}{|l|}{}                               & \textbf{Localised}    &     98.05          &  93.47            &    34.67                     &   100.00      &  81.55     \\ \hline 
\end{tabular}
\caption{NF-BoT-IoT-v2: Multi-class Detection}
\label{bot2}
\end{table}

As demonstrated in Table \ref{bot2}, the federated learning approach is superior in the detection of attacks available in the NF-BoT-IoT-v2 dataset with an average DR of 93.40\% using the DNN model and 94.61\% using the LSTM model. The centralised and localised learning approaches achieved 84.94\% and 81.84\% using the DNN model and 82.56\% and 81.55\% using the LSTM model, respectively. The reason for the average DR drop is only caused by a lack of recognition of reconnaissance attack samples where the centralised and localised learning methods achieved 44.46\% and 36.33\% respectively compared to 91.96\% detected by the federated learning method using the DNN model. Similarly, using the LSTM model, 39.01\% and 34.67\% reconnaissance attack samples were detected using the centralised and localised learning methods where the federated learning approach detected 92.88\%.

\begin{figure*}[h]
\begin{subfigure}{.5\textwidth}
  \centering
  \includegraphics[width=7cm, height=4cm]{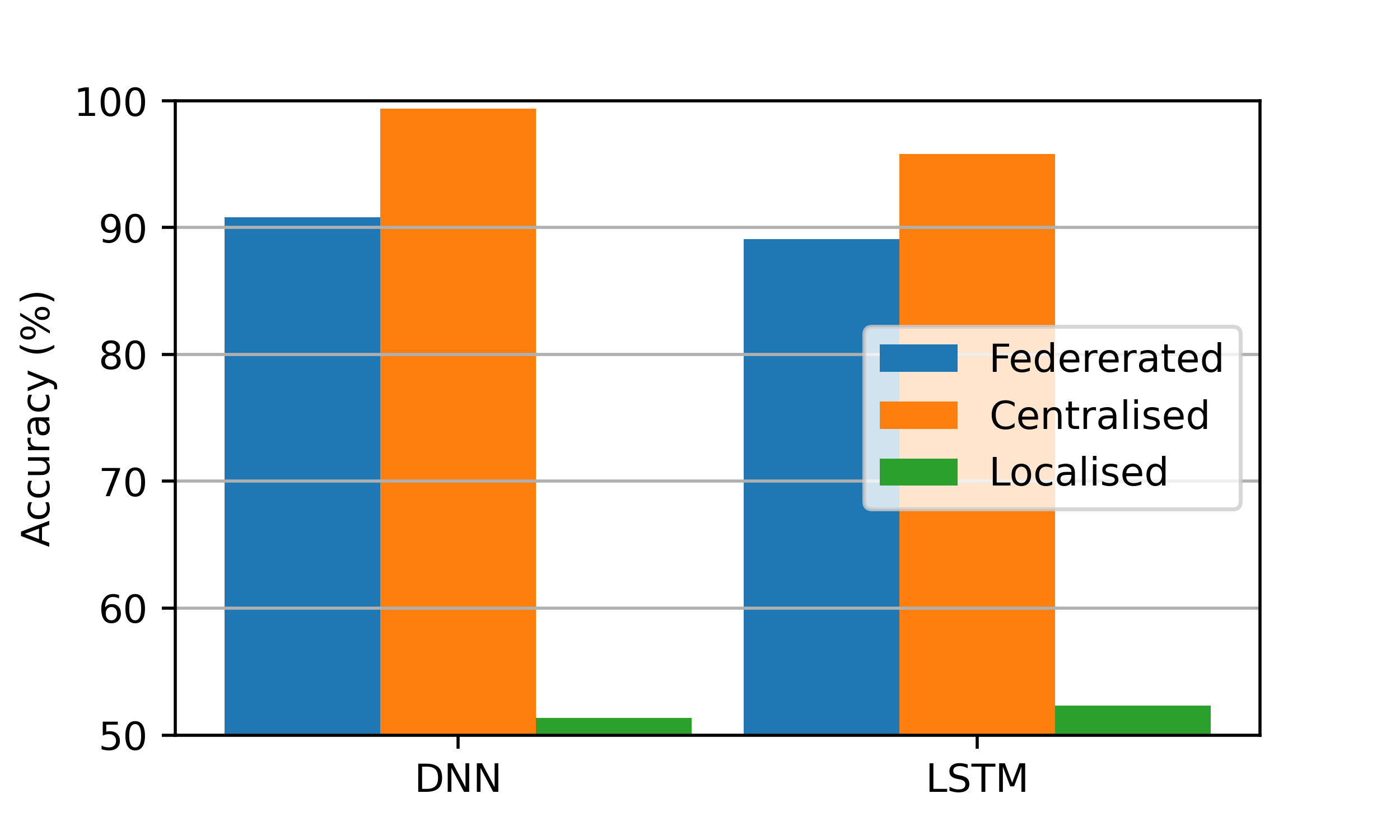} 
  \caption{NF-UNSW-NB15-v2}
  \label{k}
\end{subfigure}
\hfill
\begin{subfigure}{.5\textwidth}
  \centering
  \includegraphics[width=7cm, height=4cm]{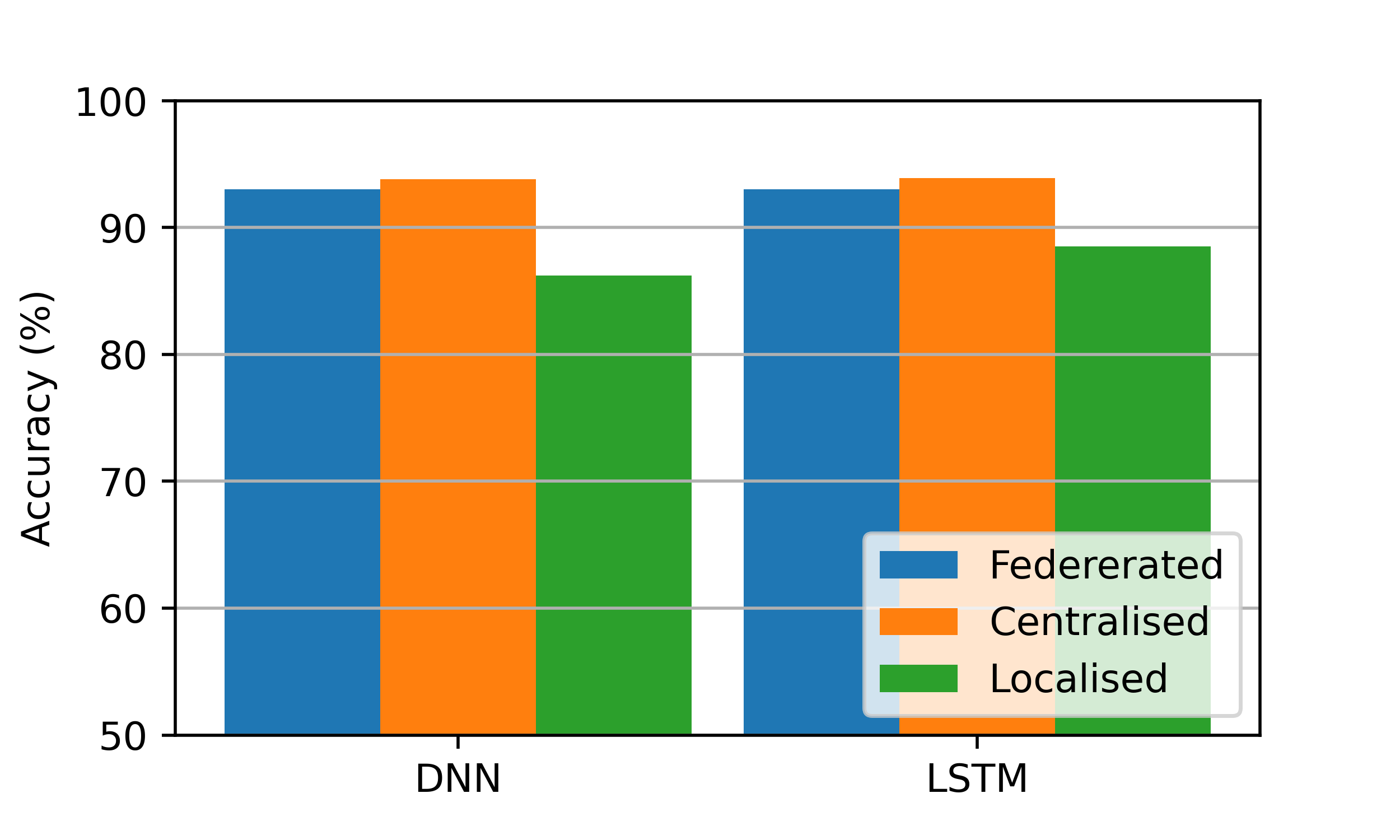}
  \caption{NF-BoT-IoT-v2}
  \label{kk}
\end{subfigure}
\caption{Binary-class Comparison}
\label{jj}
\end{figure*}

\begin{figure*}[h]
\begin{subfigure}{.5\textwidth}
  \centering
  \includegraphics[width=7cm, height=4cm]{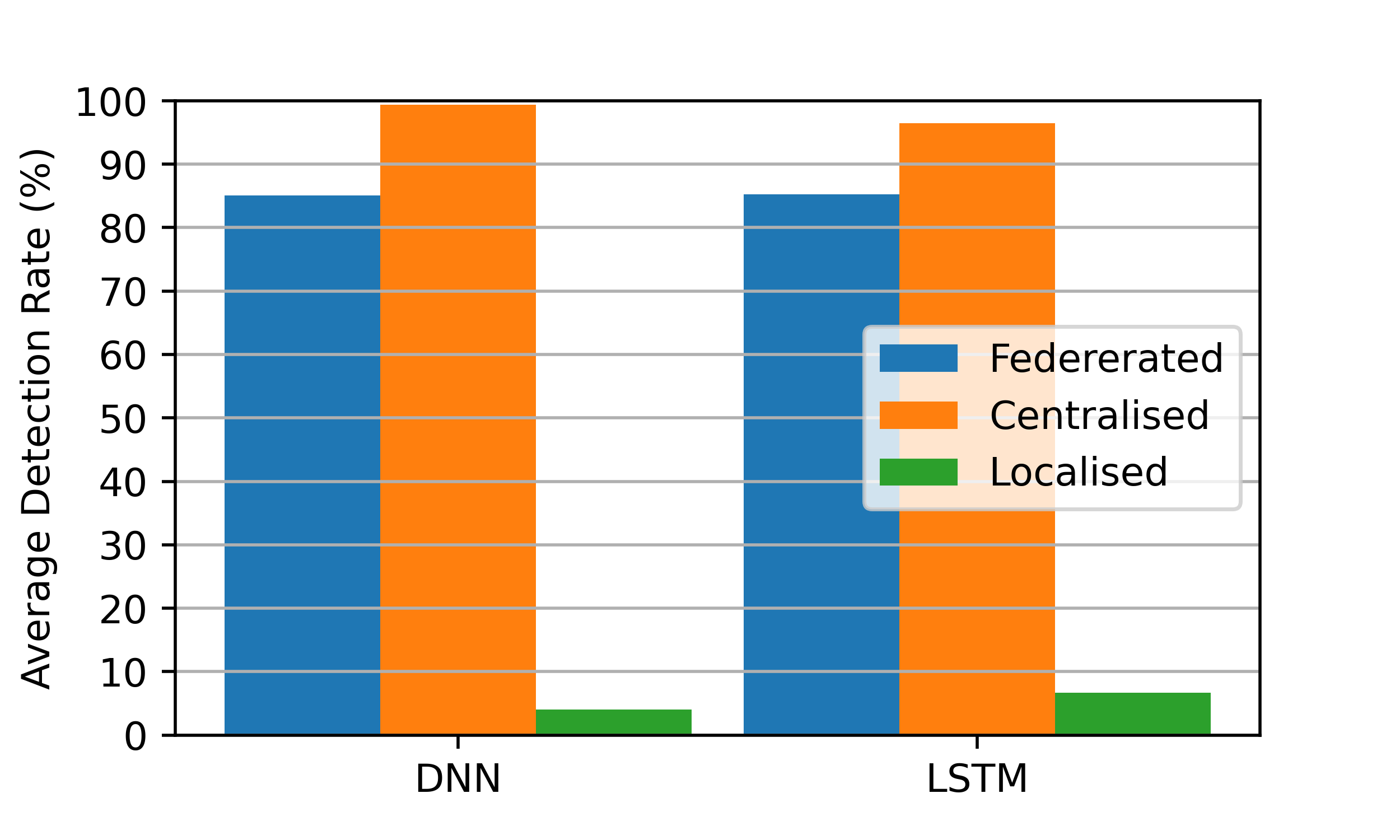} 
  \caption{NF-UNSW-NB15-v2}
  \label{l}
\end{subfigure}
\hfill
\begin{subfigure}{.5\textwidth}
  \centering
  \includegraphics[width=7cm, height=4cm]{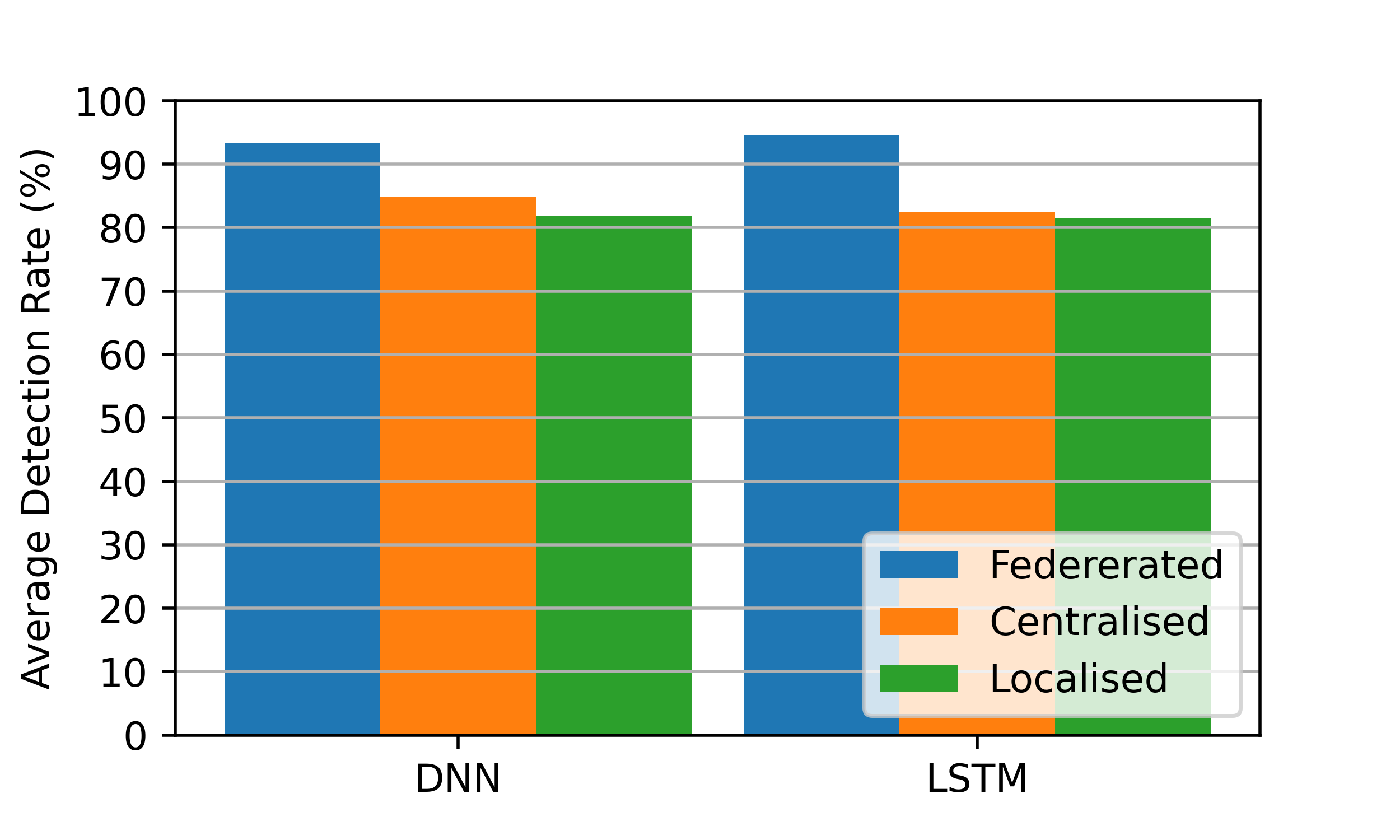}
  \caption{NF-BoT-IoT-v2}
  \label{ll}
\end{subfigure}
\caption{Multi-class Comparison}
\label{oo}
\end{figure*}

In Figures \ref{jj} and \ref{oo}, a summary of the key results is presented in bar graphs to compare the binary- and multi-classes detection results following each ML scenario. In Figure \ref{jj}, the accuracy evaluation metric is used to compare the three methods where the centralised learning method achieved the best performance using both ML models, followed by the federated learning method achieving a very similar overall detection performance. In a localised learning scenario, both models were able to learn useful patterns from the NF-UNSW-NB15-v2 dataset to classify most of the network traffic in the NF-BoT-IoT-v2 dataset. However, this was not the same case for the other way around where both models failed to achieve a reliable detection performance. In Figure \ref{oo}, the average attack DR is displayed on the y-axis, where the centralised learning and federated learning approaches were the most effective in detecting the attacks available in the NF-UNSW-NB15-v2 and NF-BoT-IoT-v2 datasets, respectively. The localised learning method did not detect most of the attacks available in the NF-UNSW-NB-v2 dataset.

Overall, a large number of experiments were conducted to evaluate and compare the performance of three ML scenarios, i.e., federated learning, centralised and localised learning. For a fair evaluation, two different ML models were used in the training and testing stages. The results demonstrate that the best set of results were often achieved by following the centralised learning approach. However, this is not possible without breaching network users' privacy by sharing sensitive data with third parties. In the real world, this might make centralised learning approaches unfeasible and costly for organisations. Therefore, the proposed scenario of a collaborative federated learning approach which achieved similar performance to the centralised approach makes it superior in terms of feasibility and preserving user privacy.

\section{Conclusion}
\label{con}
In this paper, a collaborative federated learning scheme is proposed to allow for CTI sharing between organisations to design a more effective ML-based NIDS. The collaboration between organisations attracts many benefits including the design of a robust learning model capable of detecting intrusions effectively across various organisational networks. The heterogeneity of the network data samples exposes the model to a wider variety of SOEs and attack scenarios. This reflects the real-world behaviour where each network accounts for a unique statistical distribution that ML model performance might not generalise across. The detection performance of the models is compared to centralised and localised learning scenarios. The results demonstrate that the federated learning performance is superior to the localised use case and similar to the centralised use case. However, the centralised method can not be used without breaching data privacy and security which renders it unfeasible in the real world. Future works involve enhancing the federated learning averaging algorithm to improve the detection performance of ML-based NIDS.

\bibliography{main.bib}

\end{document}